\documentclass{article}
\usepackage{spconf,amsmath,graphicx}
\usepackage{hyperref}       %
\usepackage{url}            %
\usepackage{booktabs}       %
\usepackage{amsfonts}       %
\usepackage{microtype}      %
\usepackage{eqnarray,amsmath}
\usepackage{subcaption}
\usepackage{xcolor}
\usepackage{caption}
\usepackage{setspace}
\usepackage{adjustbox}
\usepackage{multirow}

\usepackage{braket}
\usepackage{qcircuit}

\usepackage{microtype}

\title{When BERT Meets Quantum Temporal Convolution Learning \\ for Text Classification in Heterogeneous Computing}

\name{Chao-Han Huck Yang$^{1, 2*}$}
\name{%
\begin{tabular}{@{}c@{}}
Chao-Han Huck Yang$^{1}$ \qquad
Jun Qi$^{1}$ \qquad
Samuel Yen-Chi Chen$^{2}$ \\
Yu Tsao$^{3}$ \qquad
Pin-Yu Chen$^{4}$ 
\end{tabular}}
\address{$^1$ Georgia Institute of Technology, GA, USA\\$^2$Brookhaven National Laboratory, NY, USA and $^3$Academia Sinica, Taipei, Taiwan \\$^4$IBM Research, Yorktown Heights, NY, USA}

\begin{document}
\ninept
\maketitle
\begin{abstract}
The rapid development of quantum computing has demonstrated many unique characteristics of quantum advantages, such as richer feature representation and more secured protection on model parameters. This work proposes a vertical federated learning architecture based on variational quantum circuits to demonstrate the competitive performance of a quantum-enhanced pre-trained BERT model for text classification. In particular, our proposed hybrid classical-quantum model consists of a novel random quantum temporal convolution (QTC) learning framework replacing some layers in the BERT-based decoder. Our experiments on intent classification show that our proposed BERT-QTC model attains competitive experimental results in the Snips and ATIS spoken language datasets. Particularly, the BERT-QTC boosts the performance of the existing quantum circuit-based language model in two text classification datasets by 1.57\% and 1.52\% relative improvements. Furthermore, BERT-QTC can be feasibly deployed on both existing commercial-accessible quantum computation hardware and CPU-based interface for ensuring data isolation.

\end{abstract}


\section{Introduction}
Text classification (e.g., intent detection) from human spoken utterances~\cite{stolcke2000dialogue, wang2006computational} is an essential element of a spoken language understanding~\cite{wang2005spoken} (SLU) system. Learning intent information often involves various applications, such as on-device voice assistants~\cite{coucke2018snips} and airplane travel information systems~\cite{hemphill1990atis}.
Recently, Bidirectional Encoder Representations from Transformers~\cite{devlin2019bert} (BERT) has caused a stir in the lexical modeling community by providing competitive results in intent classification as a common SLU application that we torched upon in this work.
However, recent works~\cite{carlini2019secret,carlini2020extracting} have raised new concerns about data leakage from large-scale language models such as BERT, which can involve sensitive information like personal identification. New regulation requirements~\cite{voigt2017eu} (e.g., GDPR) on data protection, privacy~\cite{yang2021pate}, and security can further motivate advanced investigations in designing distributed algorithms on heterogeneous computing devices.   

In recent two years, cloud-accessible quantum devices (e.g., IBM-Q) have shown unique characteristics and empirical advantages in many applications~\cite{yang2020decentralizing, arute2019quantum}, such as model compression, parameter isolation, and encryption. A noisy-intermediate-scale-quantum (NISQ) device is a main hardware category that empowers the quantum advantages by using only a few number of qubits (5 to 100). When the limitation of quantum resource is concerned, variational quantum circuit (VQC) has been studied on the design of quantum machine learning like quantum support vector machine, where the VQC algorithm requires no quantum error correction as a matching working on NISQ devices. Projecting classical data into high-dimensional quantum feature space has been proven its quantum advantages~\cite{liu2021rigorous, huang2021power} (e.g., a quadratic speed-up~\cite{liu2021rigorous} and better presentation power~\cite{yang2020decentralizing}) on some classifications tasks. For example,
Yang \emph{et al.}~\cite{yang2020decentralizing} demonstrates the quantum advantages on a speech command recognition system by applying the VQC algorithm to extract acoustic features for representation learning with a vertical federated learning architecture. The proposed vertical federated learning could benefit from quantum advantages\footnote{Quantum advantage means that a programmable quantum device~\cite{boixo2018characterizing} can solve a specific problem that is intractable on classical computers.}, such as parameter isolation with random circuit learning, for the acoustic modeling task. However, designing an end-to-end neural model on SLU tasks is still an open problem with a potential impact on heterogeneous computing devices.

Motivated by the very recent success \cite{yang2020decentralizing} in the quantum circuit-based acoustic modeling, we put forth a new gradient-based training to include pre-trained BERT models and enhance its data protection by leveraging upon VQC learning in this work. To strengthen the parameter isolation on BERT, our work proposes a novel hybrid classical-quantum system with quantum temporal convolution (QTC) learning, which is composed of a pre-trained BERT model (on a classical computer) and a new type of one-dimensional random circuit~\cite{henderson2020quanvolutional, yang2020decentralizing} (on a quantum device). In comparison with the benchmark pre-trained BERT model, our proposed architecture can maintain competitive experimental performance considering the SLU benchmark solution running in homogeneous machines. Notably, in this work, compared with the classical deep neural network (DNN) models, the use of QTC lowers the model complexity. The VQC design of QTC is to use the quantum circuit as a temporal convolutional features projector on the text embeddings that requires only a few qubits (e.g., 4 to 9) to set up our quantum platform. 

\begin{figure}[htbp]
\centering \includegraphics[width=0.75\linewidth]{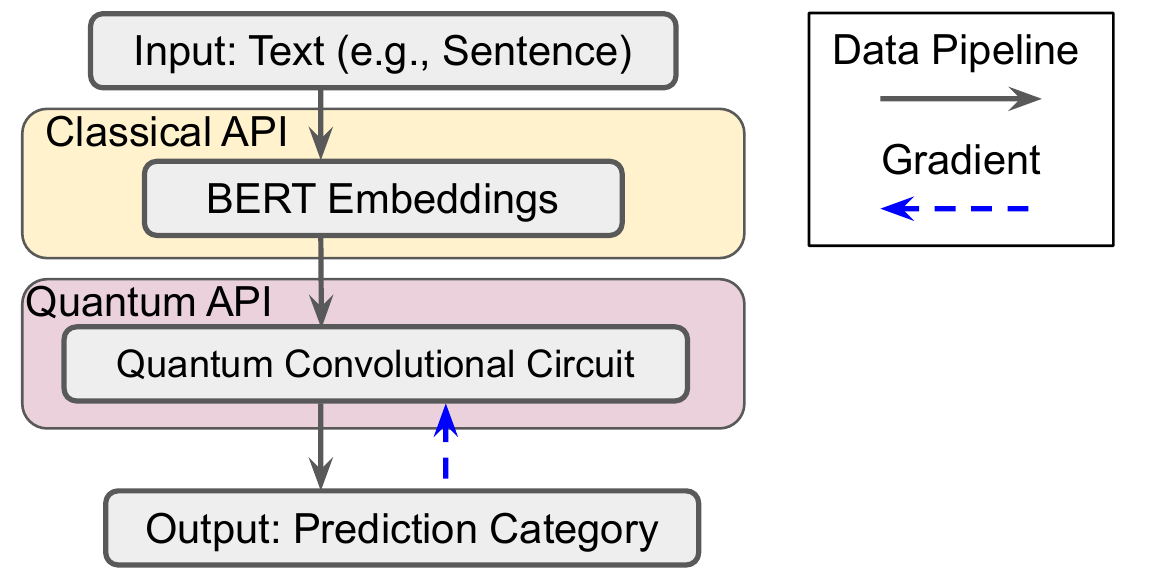}
\caption{BERT-QTC for vertical federated~\cite{yang2020decentralizing} text classification.}
\label{fig:sys}
\end{figure}

As shown in Figure~\ref{fig:sys}, the design of such a BERT-QTC model is in a regime of hybrid classical-quantum architecture, where the word embedding can be offline attained by utilizing DNN models on classical computers before going through quantum devices. Table \ref{tab:overview} shows an overview of classical, quantum~\cite{liu2013novel}, and heterogeneous approaches for SLU. To the best of our knowledge, this is the \textbf{first} work to investigate a BERT-based hybrid classical-quantum model for text classification (e.g., intent detection) task with competitive empirical results on quantum circuit learning on the NISQ device.

\begin{table}[h]\footnotesize
\centering
\vspace{-2mm}
\caption{An overview of different natural language processing (NLP) approaches: neural language model (NLM), logic programming (LP), and its quantum variants. Our work belongs to hybrid CQ to access quantum advantages (QA).}
\label{tab:overview}
\begin{adjustbox}{width=0.48\textwidth}
\begin{tabular}{|l|l|l|l|l|}
\hline
Approach & Input &  Model & Output & Challenges \\ \hline
Classical & bits & NLM \& LP & bits & data protection \\ \hline
Quantum & qubits & Quantum LP~\cite{coecke2020foundations} & qubits & hardware limits\\ \hline
hybrid CQ & bits & NLM + VQC & bits &  model design \\ \hline
\end{tabular}
\end{adjustbox}
\vspace{-4mm}
\end{table}

\section{Related Work}
\subsection{Quantum Algorithms for Language Processing}
Recent years have witnessed a rapid development~\cite{havlivcek2019supervised} of quantum machine learning on near-term quantum devices (5 to 200 qubits). Encoding classical data into high-dimensional quantum features has proven rigorous quantum advantages on classification tasks with recent theoretical justification~\cite{liu2021rigorous, huang2021power}. In particular, VQC learning~\cite{mitarai2018quantum} is utilized as parametric models to build a QNN model. The research of VQC is of significance in itself because they constitute the setting of quantum computational experiments on NISQ devices.
Although business-accessible quantum services (e.g., IBM-Q and Amazon Braket) are still actively being developed, several works have attempted to incorporate quantum algorithms for language processing tasks. For example, \cite{liu2013novel} Liu \emph{et al.} consider a novel classifier as the physical evolution process and described the process with quantum mechanical equations. In \cite{sordoni2013modeling, basile2017towards}, the theory of quantum probability is proposed to build a general language model for information retrieval.

Moreover, Blacoe \emph{et al.} \cite{blacoe2013quantum} leverage upon quantum superposition and entanglement to investigate the potential of quantum theory as a practical framework to capture the lexical meaning and model semantic processes such as word similarity and association. However, those works aim at the use of quantum concepts to build quantum language models in theory, and they are different from the recent quantum circuit architectures featured with small-to-none quantum error correction in the NISQ devices.  

Recently, Meichanetzidis \emph{et al.}\cite{meichanetzidis2020quantum} first investigated computational logic relationships for encoding graphs into circuit learning on NISQ computers. Coecke \emph{et al.} ~\cite{coecke2020foundations} investigate theoretical foundation and advantages of encryption on using the VQC learning on linguistic modeling. In particular, their quantum circuits are employed as a pipeline to map semantic and grammar diagrams and pave the way to near-term applications of quantum computation devices. Lorenz \emph{et al.}~\cite{lorenz2021qnlp} introduced a concept of compositional semantic information for text classification tasks on the quantum hardware by simply using a mapping from lexical segments to a quantum circuit.  Nevertheless, beyond grammar modeling, how to design a specific quantum model is still a wide-open topic with potential impacts on ``large-scale'' data (e.g., spoken language processing and applications).

\subsection{BERT and Heterogeneous Computing}

Devlin \emph{et al.} \cite{devlin2018bert} firstly put forth the architecture of BERT, which mainly applies the masking and bidirectional training of Transformer equipped with an attention model to language modeling. BERTs~\cite{lan2019albert} can be used for a wide variety of language tasks since different SLU tasks only require a few more layers to the core model for fine-tuning. When the training time and data protection become significant concerns for BERTs, heterogeneous computing architectures, such as distributed training and federated learning~\cite{chen2021federated}, provide a new perspective to ensure data isolation as protection and improve training efficiency.  

In our previous work~\cite{yang2020decentralizing}, we propose a new vertical federated learning architecture that combines new convolution-encoded randomized VQC and recurrent neural network (RNN) (denoted as QCNN-RNN) for acoustic modeling preserving quantum advantages. The QCNN-RNN system first uses a quantum convolution embedding layer to encode features in high-dimensional Hilbert space, then encode quantum features in a classical format for the representation learning with RNN. Different with the quantum acoustic modeling task, we want to advance a new VQC design working on the SLU task considering the benchmark pre-trained language model (e.g., BERT) in this work.

\section{Method}
\label{sec:length}

\subsection{BERT for Text Embedding}

\begin{figure}[htbp]
\centering \includegraphics[width=0.95\linewidth]{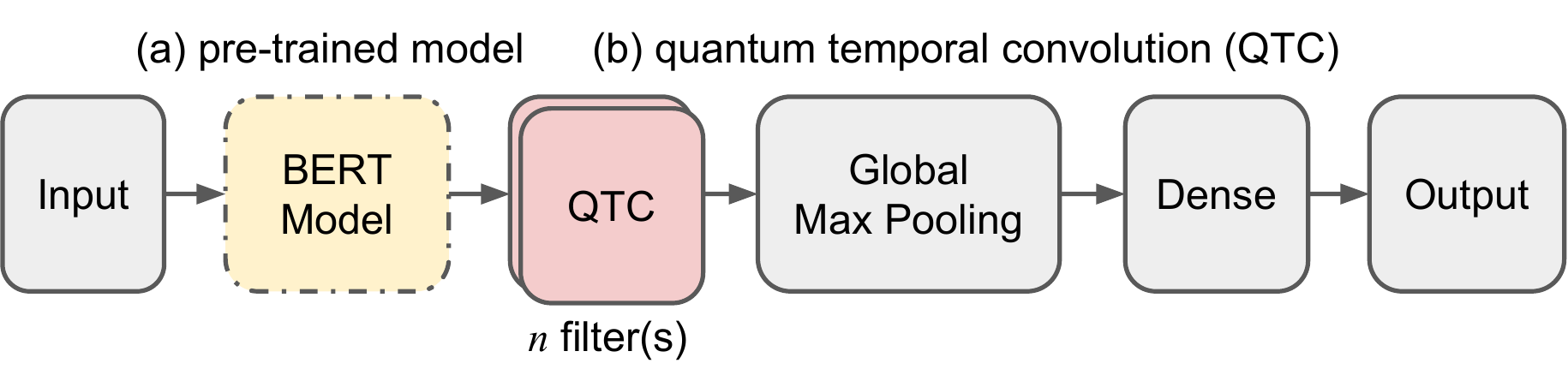}
\caption{The BERT-QCC comprises with (a) BERT and (b) quantum temporal convolutional (QTC) learning with random circuits. }
\label{fig:qnn}
\end{figure}
We first use BERT as a pre-trained language model to extract hidden representation features from text sequences as shown in Figure~\ref{fig:qnn} (a). The BERT based language models consist of a multi-layer bidirectional Transformer-based encoder, where the input corresponds to a combination of WordPiece embeddings, positional embeddings, and the segment embedding. Symbolically, given an input token sequence $\boldsymbol{x}=\left(x_{1}, \ldots, x_{T}\right)$
the BERT model outputs are computed by:
\begin{equation}
\mathbf{H}=\mathbf{BERT}\left(x_{1}, \ldots, x_{T}\right),
\label{eq:bert}
\end{equation}
where $\mathbf{H}=\left(\boldsymbol{h}_{1}, \ldots, \boldsymbol{h}_{T}\right)$ and $\boldsymbol{h}_{i}$ is contextual semantic representation embedding of each token in Eq. (\ref{eq:bert}).

\textbf{Neural Network Intent Decoder.}
We select a hidden state of the first special token for intent classification after a dropout layer to avoid over-fitting. Conventionally, the hidden state is encoded by a neural network  function ($\boldsymbol{f}(\mathbf{H};\theta_{\boldsymbol{f}})$) for extracting sentence-level semantic representation. Finally, an unknown intent can be predicted as:
\begin{equation}
y^{i}=\operatorname{softmax}\left(\mathbf{W}^{i} \boldsymbol{f}(\mathbf{H};\theta_{\boldsymbol{f}})+\boldsymbol{b}^{i}\right),
\label{eq:2:intent}
\end{equation}
where $y^{i}$ is a predicted intent label for a test sequence. 
The model is end-to-end trained by minimizing the cross-entropy loss between predicted text labels and their corresponding ground-truth labels. 
\subsection{Quantum Temporal Convolution with Random Circuit}
The expressive power of quantum convolution has been recently studied in computer vision~\cite{henderson2020quanvolutional} and speech processing~\cite{yang2020decentralizing, qi2021qtn} by using quantum space encoding to project classical data into rich representations of features for classification. Deploying a quantum circuit with randomized assigned (non-trainable) parameters could beat both trainable and non-trainable convolutional filters in the previous study~\cite{yang2020decentralizing}. We further consider a new hybrid classical-quantum decoder design, which replace the perception network ($\boldsymbol{f}$) in Eq. (\ref{eq:2:intent}) with a randomized quantum circuit decoder as shown in Fig.~\ref{fig:qnn} (b).  We use a max length of 50 for pre-trained BERT models.

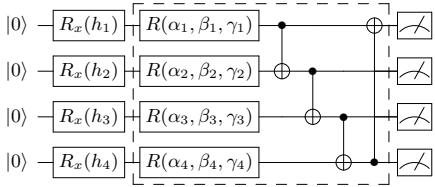
\begin{figure}[htbp]
\begin{center}
\scalebox{0.8}{
\begin{minipage}{10cm}
\Qcircuit @C=0.8em @R=0.8em {
\lstick{\ket{0}} & \gate{R_x(h_1)} & \gate{R(\alpha_1, \beta_1, \gamma_1)} & \ctrl{1}   & \qw       & \qw      & \targ    & \meter \qw \\
\lstick{\ket{0}} & \gate{R_x(h_2)} & \gate{R(\alpha_2, \beta_2, \gamma_2)} & \targ      & \ctrl{1}  & \qw      & \qw     &  \meter \qw \\
\lstick{\ket{0}} & \gate{R_x(h_3)} & \gate{R(\alpha_3, \beta_3, \gamma_3)} & \qw        & \targ     & \ctrl{1} & \qw     &  \meter \qw \\
\lstick{\ket{0}} & \gate{R_x(h_4)} & \gate{R(\alpha_4, \beta_4, \gamma_4)} &\qw        & \qw       & \targ    & \ctrl{-3}&  \meter \qw \gategroup{1}{3}{4}{7}{.7em}{--}\qw 
}
\end{minipage}
}
\end{center}
\caption[Variational quantum circuit component]{{\bfseries Deployed quantum circuit.} The VQC component contains three major parts: encoding, learning and quantum measurement. Here we use the angle encoding scheme to encode the input values $h_1 \cdots h_4$  (latent embedding; taking four qubits, for example) by treating them as rotation angles along $x$-axis. The learning part uses general unitary rotation $R$. There are three parameters $\alpha$, $\beta$ and $\gamma$ in each $R$. The controlled-NOT (CNOT) gates are used to entangle quantum states from each qubit. The final quantum measurement part will output the Pauli-$Z$ expectation values of each qubit. }

\label{Fig:Basic_VQC_1}
\end{figure}

\begin{figure}[htbp]
\centering \includegraphics[width=0.85\linewidth]{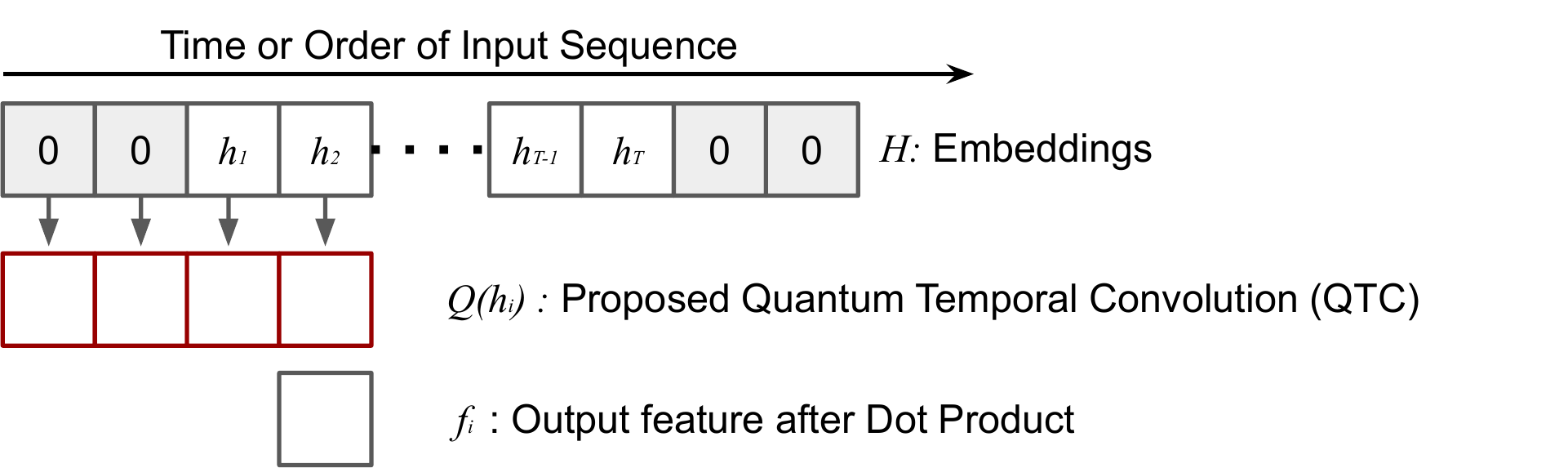}
\caption{Computing framework for quantum temporal convolution (QTC) with random variational circuit from input text embeddings. }
\label{fig:1ddc}
\end{figure}

\textbf{Variational Quantum Circuit Decoder.} We adopt the concept from variational quantum circuit learning and define a latent encoding function $\hat{\boldsymbol{f}}$, where it contains angle encoding, parameterized rotation, and quantum-to-classical state decoding. We use the VQC as shown in Fig.~{\ref{Fig:Basic_VQC_1}}, made by strong entanglement~\cite{chen2020variational} circuit.  First, the Bert embeddings $[\mathbf{h}_1, \cdots, \mathbf{h}_T]$ are feeding into a sliding window for the temporal quantum convolutional filter (e.g., coordinated with time or order of input), where we take window size $I=4$ for example as one desired qubits size shown in Fig.~\ref{fig:1ddc}. We then extract latent embedding $[h_1, h_2, h_3, h_4]$ as inputs to the VQC layer.

The latent features ($H$) are the input of a quantum circuit layer, $\mathbf{Q}$, that learns and encodes latent sequence:
\begin{equation}
    \hat{\boldsymbol{f}}=\mathbf{Q}(H; \mathbf{e}, \theta_{\mathbf{q}}, \mathbf{d})
    \label{eq:1:vqc}
\end{equation}
In Eq. (\ref{eq:1:vqc}), the approximation process of a quantum circuit block, $\mathbf{Q}$, depends on the encoding initialization $\mathbf{e}$, the quantum circuit parameters, $\theta_{\mathbf{q}}$, and the decoding measurement $\mathbf{d}$. When we stack more VQC blocks, we expect that a gradient vanishing issue would occur and degrade the representation learned from a hybrid QNN model whose parameters are trained by joint DNN-VQC gradient updates.  

\begin{table}[ht!]
\centering
\caption{Mathematical notation for VQC learning.}
\label{tab:notation}
\begin{tabular}{@{}l|l@{}}
\toprule
Symbol                                                                                                    & Meaning                                       \\ \midrule
$|0\rangle$                                 &  quantum state $0$           \\
R$_x$                                       & rotation gate along $x$-axis  \\
$\theta_{\mathbf{q}}$=$\{\alpha, \beta, \gamma\}$                                                                               & random parameters                             \\
CNOT                                        & controlled-NOT gate        \\

\bottomrule
\end{tabular}
\end{table}

In this VQC, the elements $h_1 \cdots h_4$ of the input vectors are used as rotational angles for single-qubit quantum rotation gate $R_x$ on each qubit to encode the classical values. After encoded into a quantum state, the state is processed through a \emph{learnable} or \emph{optimizable} layer (in grouped box). The parameters labeled with $\alpha$, $\beta$, and $\gamma$ are for the optimization. The CNOT gates are applied to entangle the qubits. In the final step, the quantum measurement procedure is carried out. In this part, we retrieve the Pauli-$Z$ expectation values for each qubit. The obtained result is a $4$-dimensional vector which can be further processed by other classical or quantum routines.

Figure~\ref{Fig:Basic_VQC_1} summarize characterises of the variational quantum circuit, where there are $4$ quantum channels and each channel is mutually entangled with each other by applying the CNOT gate. Given the unitary matrix $\mathbb{U}$ representing the multiplication of all quantum gates and taking $B$ as observable matrix, we attain that:

\begin{equation}
    \hat{\boldsymbol{f}}(h; \boldsymbol{\theta_{\text{random init}}}) = \langle h | U^{\dagger}(\boldsymbol{\theta}_{i}) B U(\boldsymbol{\theta}_{i})  | h \rangle.
    \label{eq:gate}
\end{equation}

The hyper-parameters ($\boldsymbol{\theta_{\text{random init}}}$) for rotation gates in Eq. (\ref{eq:gate}) are randomly initialized and not considered to be updated during the training phase, which is aimed to provide parameters protection in heterogeneous computing architectures (or simulation API) against model inversion attacks~\cite{fredrikson2015model} and parameters leakages~\cite{duc2014unifying}. As shown in Fig.~\ref{fig:qnn}, the temporal quantum convolution could be applied with $n$ filters to map features before computing with a global max-pooling layer. As \textbf{the first attempt} to construct quantum temporal convolution, we select a filters number from $\{1,2,3,4\}$ under the 9 qubits requirement for common commercial quantum hardware (e.g., IBMQ).

To evaluate the effectiveness of proposed QTC architectures, we select three additional ``random'' encoder baselines (similar to QTC) with text embeddings for text classification: (1) BERT with a random temporal convolutional network (TCN); (2) word2vec~\cite{mikolov2013efficient} with random TCN; (3) word2vec with random QTC, all followed the same filter numbers and global max-pooling after the deployed random encoders.

We will study how random QTC learning benefits BERT-based heterogeneous architecture in the experimental section. From a system-level perspective, the proposed QTC learning reduces the risk of parameter leakage~\cite{duc2014unifying, leroy2019federated, chen2019federated} from \textbf{inference-time} attackers, and it tackles data isolation issues with its architecture-wise advantages~\cite{dwork2015reusable} on encryption~\cite{yao1993quantum} and the feature of without accessing the data directly~\cite{yang2019federated, yang2020decentralizing}.

\section{Experiment}
\subsection{Quantum Computing Hardware}
We use PennyLane as an open-source and reproducible environment, which uses differential programming to calculate gradient to update gradient circuits. A hardware back-end of PennyLane could integrate from CPU-simulation, QPU (supported by Amazon Braket), and TPU (supported by Google Floq). Since fine-tuning time of BERTs is often extensive, we first use a CPU-simulated VQC environment to train the proposed BERT-QTC model. Then, we evaluate our hybrid classical-quantum models on Amazon Braket and Floq hardware and report test accuracy by 10-fold cross-validation. Referring to the established work~\cite{yang2020decentralizing} on VQC learning, we consider a NISQ device with 4 to 16 qubits, which allows us to perform up to 9 class predictions. Furthermore, we refine our dataset according to the hardware settings, and clarify that NISQ's size constraints limit current applications of proposed BERT-QTC working toward word-level slot-filling tasks (with 120 to 72 classes). We believe that the development of commercial-accessible NISQ could gradually resolve this challenge. For vertical 
learning setting, we use secure multiparty computation (MPC) protocol~\cite{cramer2015secure} between local instance and quantum device during the virtualization. 

\subsection{Dataset and Setup}

\textbf{Snips Dataset:} 
To evaluate the proposed framework in complex spoken language corpora, we select Snips~\cite{coucke2018snips} dataset. Snips dataset is a collection of spoken utterances from a personal voice assistant, whose domains cover speech command, music searching, and smart home request. 
The training set includes 13,084 utterances, and the test set includes 700 utterances. We use another 700 utterances as the development set. There are 7 types of intent classes, where the number of samples for each intent is nearly the same.

\begin{figure}[htbp]
\centering \includegraphics[width=0.60\linewidth]{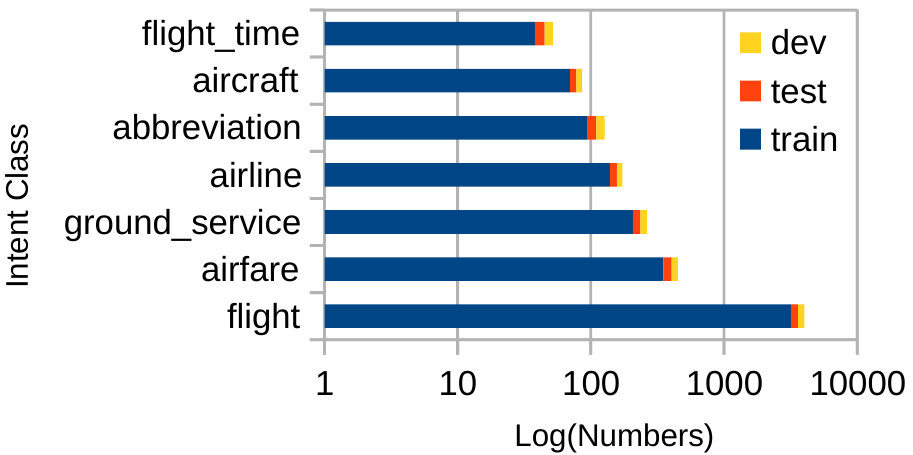}
\caption{ATIS$_7$ Dataset for BERT-QTC experiments.}
\label{fig:atis}
\end{figure}
\vspace{-2mm}

\textbf{ATIS Dataset:} The ATIS (Airline Travel Information System) dataset~\cite{hemphill1990atis} is widely used in NLU research containing spoken corpus, including intent classification and slot filling labels for flights reservation. Since the language scripts from ATIS are collected from human speakers, ATIS is selected to highlight the data sensitivity under GDPR policy~\cite{voigt2017eu}. To compare with Snips as a slightly unbalanced classification study, we further select a subset of ATIS containing top-7 classes (denoted as ATIS$_7$) of original ATIS. As shown in Fig.~\ref{fig:atis}, the deployed subset include 90.93\% of the original data, where training, development, and testing sets contain 4072, 545, 556 utterances, respectively. As a remark, we have conducted text classification experiments with a full ATIS dataset, but the results have a large variance ($\pm$ 7.32\%) on both random TCN and QRC encoders due to unbalanced intent labels in a long-tail distribution~\cite{yang2021multi}.

\textbf{BERT Model:} We use an open source BERT~\cite{devlin2018bert} model to connect the QTC encoder. The pre-trained models of BERT are frozen in the heterogeneous computing process as a universal encoder to take sentences into embedding. We use the official pretrained BERT (large) model from TensorFlow Hub with 1024-hidden dimensions, 24 layers, and 12 attentions heads with 330M pre-trained parameters. 

\subsection{BERT-QTC Performance}
Since we are working on sentence-level text classification (intent as labels) on the two deployed datasets, we first select two different text embeddings methods from pre-trained word2vec and BERT embeddings to compared the proposed random QTC encoder with a random TCN encoder. Furthermore, we aim to study different hyper-parameters setup for TCN and QTC architecture. As a remark, both TCN and QTC could be simulated with CPU-based environments or API, where QTC is featured with better representation mapping with theoretical justifications and an option running with quantum hardware to preserve full quantum advantages.

As shown in Tab.~\ref{tab:snip1} (Snips) and ~\ref{tab:atis1} (ATIS$_7$), the BERT-QTC model in vertical federated architecture performs the best average prediction accuracy performance, which attains $\mathbf{96.62}$\% in Snips and $\mathbf{96.98}$\% in ATIS$_7$ compared with TCN based architectures and word2vec with QTC encoder. Moreover, we investigate the convolution filter and kernel setups on BERTs-QTC models from the second, fourth row in Tab.~\ref{tab:snip1} and ~\ref{tab:atis1}. BERT-QTC demonstrates more significant improvement with its word2vec-QTC baseline when two filters have been used. Interestingly, we find out that utilize QTC encoder shows a general performance-boosting in the two deployed SLU datasets. Proposed BERT-QTC federated models perform relative improvements of \textbf{+1.57\%} in Snips and \textbf{+1.52\%} in ATIS$_7$ compared with other heterogeneous computing ablations. The statistical variances are $\leq 0.52\%$ in our experiments. 

\begin{table}[ht!]
\centering
\caption{Average accuracy on intent classification for Snips with a set of different number (n) of convolutional filter and kernel size (k).}
\label{tab:snip1}
\begin{adjustbox}{width=0.48\textwidth}
\begin{tabular}{ccccccccc}
\toprule
\multicolumn{1}{c}{Embedding} & \multicolumn{4}{c}{word2vec} & \multicolumn{4}{c}{BERT} \\ \midrule \midrule
(n,k) & (1,4) & (2,2) & (2,3) & (2,4) & (1,4) & (2,2) & (2,3) & (2,4) \\ \midrule
TCN & 82.02 & 83.37 & 82.90 & 83.15 & 95.48 & 95.23 & 95.12 & 95.27 \\ \midrule
QTC & \textbf{83.32} & \textbf{83.94 }& \textbf{83.61} & $\mathbf{84.64}$ & \textbf{96.41} & \textbf{96.42} & $\mathbf{96.62}$ & \textbf{96.44} \\ \bottomrule
\end{tabular}
\end{adjustbox}
\end{table}
\vspace{-4mm}

\begin{table}[ht!]
\centering
\caption{Average accuracy on intent classification for ATIS$_7$ with a set of different number (n) of convolutional filter and kernel size (k).}
\label{tab:atis1}
\begin{adjustbox}{width=0.48\textwidth}
\begin{tabular}{ccccccccc}
\toprule
\multicolumn{1}{c}{Embedding} & \multicolumn{4}{c}{word2vec} & \multicolumn{4}{c}{BERT} \\ \midrule \midrule
(n,k) & (1,4) & (2,2) & (2,3) & (2,4) & (1,4) & (2,2) & (2,3) & (2,4) \\ \midrule
TCN & 80.09 & 80.22 & 80.91 & 82.34 & 95.18 & 95.03 & 94.95 & 95.23 \\ \midrule
QTC & \textbf{81.42} & \textbf{82.49 }& \textbf{83.82} & $\mathbf{83.95}$ & \textbf{96.69} & \textbf{96.92} & \textbf{96.32} & $\mathbf{96.98}$ \\ \bottomrule
\end{tabular}
\end{adjustbox}
\end{table}
\vspace{-2mm}

\section{Conclusion}
In this paper, we propose a novel hybrid classical-quantum architecture to strengthen the BERT model with a quantum circuit decoder via \textbf{quantum temporal convolution} (QTC) with random circuit learning. The proposed BERT-QTC models show competitive results for text classification as one prompted finding for standard SLU tasks. Moreover, our VQC encoders are capable of deploying on both existing quantum hardware, and the simulator requires only a small amount of qubits (4 to 8 qubits). The proposed QTC can enhance the data protection on top of BERT models in the vertical federated learning setting. 

\clearpage
\begin{spacing}{0.6}
\footnotesize
\bibliographystyle{IEEEtrans}
\bibliography{anthology,acl2021,vqc}
\end{spacing}
\clearpage

\end{document}